\documentclass[runningheads]{llncs}
\usepackage{graphicx}
\usepackage{tikz}
\usepackage{comment}
\usepackage{amsmath,amssymb} % define this before the line numbering.
\usepackage{color}
\usepackage{booktabs}
\usepackage{multirow}
\usepackage{colortbl}
\definecolor{mygray}{gray}{.85}
\usepackage[accsupp]{axessibility}  % Improves PDF readability for those with disabilities.
\usepackage{float}

\begin{document}

\pagestyle{headings}
\mainmatter
\def\ECCVSubNumber{4278}  

\title{Multimodal Transformer with Variable-length Memory for Vision-and-Language Navigation} % Replace with your title

% INITIAL SUBMISSION 
\begin{comment}
\titlerunning{ECCV-22 submission ID \ECCVSubNumber} 
\authorrunning{ECCV-22 submission ID \ECCVSubNumber} 
\author{Anonymous ECCV submission}
\institute{Paper ID \ECCVSubNumber}
\end{comment}
%******************

% CAMERA READY SUBMISSION
%\begin{comment}
\titlerunning{MTVM}
% If the paper title is too long for the running head, you can set
% an abbreviated paper title here
%
\author{Chuang Lin\inst{1} \thanks{This work was performed while Chuang Lin worked as an intern at
ByteDance.} \and
Yi Jiang\inst{2} \and
Jianfei Cai\inst{1} \and 
Lizhen Qu\inst{1} \and
Gholamreza Haffari\inst{1} \and
Zehuan Yuan\inst{2} }
\authorrunning{C. Lin et al.}
% First names are abbreviated in the running head.
% If there are more than two authors, 'et al.' is used.
%
\institute{\mbox{Monash University \and
 ByteDance}
}
%\end{comment}
%******************
\maketitle

\begin{abstract}
Vision-and-Language Navigation (VLN) is a task that an agent is required to follow a language instruction to navigate to the goal position, which relies on the ongoing interactions with the environment during moving.
Recent Transformer-based VLN methods have made great progress benefiting from the direct connections between visual observations and language instructions via the multimodal cross-attention mechanism.
However, these methods usually represent temporal context as a fixed-length vector by using an LSTM decoder or using manually designed hidden states to build a recurrent Transformer. Considering a single fixed-length vector is often insufficient to capture long-term temporal context, 
in this paper, we introduce Multimodal Transformer with Variable-length Memory (MTVM) for visually-grounded natural language navigation by modeling the temporal context explicitly. 
Specifically, MTVM enables the agent to keep track of the navigation trajectory by directly storing activations in the previous time step in a memory bank.
To further boost the performance, we propose a memory-aware consistency loss to help learn a better joint representation of temporal context with random masked instructions.
We evaluate MTVM on popular R2R and CVDN datasets.
Our model improves Success Rate on R2R test set by 2\% and reduces Goal Process by 1.5m on CVDN test set.
Code is available at: \url{https://github.com/clin1223/MTVM}.

\end{abstract}

\section{Introduction}

\begin{figure}[thbp]
\includegraphics[width=1.0\linewidth]{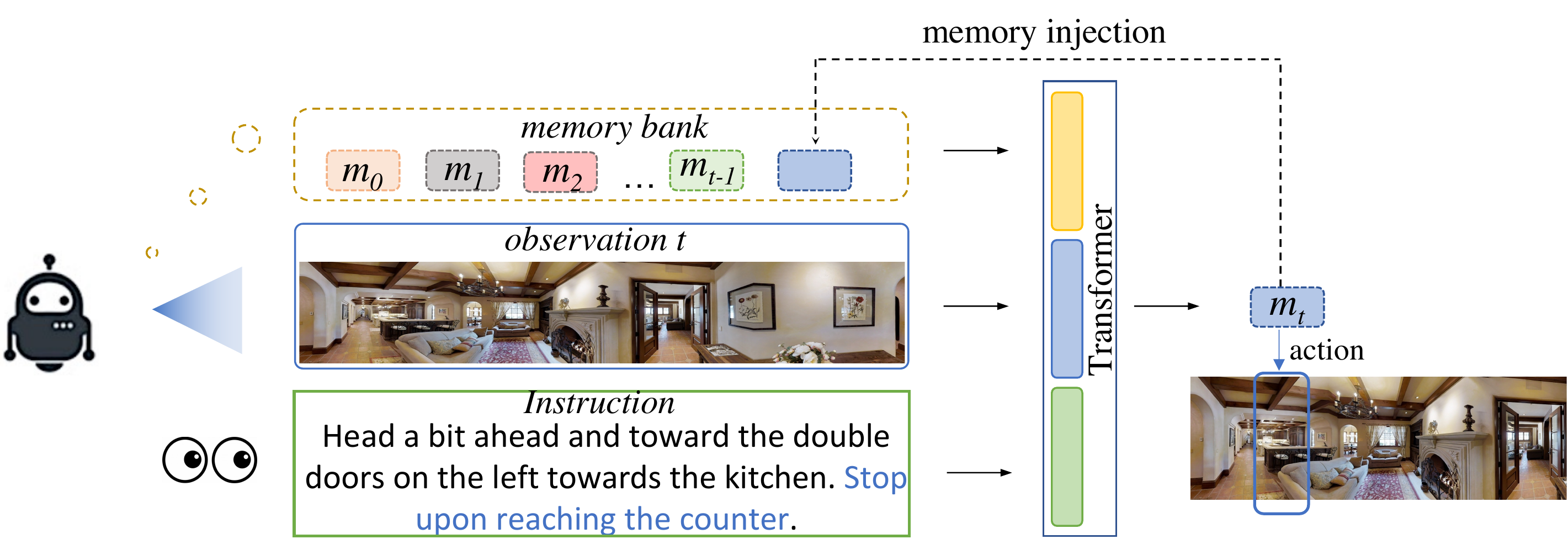}
\caption{In contrast to most existing methods that utilize a fixed-length vector to represent temporal context, we equip the agent with the capability to model long-term dependency. At each step \textit{t}, MTVM takes all the tokens stored in the memory bank as the temporal context input. After making a decision, it adds a memory token $m_t$ by simply reusing the output activation corresponding to the action at step $t$.}
\label{memoried transformer}
\end{figure}

Enabling robots to assist humans in real world has been  desired so long  in AI~\cite{chen2011learning,tellex2011understanding,guadarrama2013grounding}. To achieve it, one crucial capability of robots is to be able to follow human instructions to navigate their environments. %, especially in environments that have never been seen before.
Vision-and-Language Navigation (VLN) is the task where an embodied agent is required to follow language instructions to navigate to a goal position.
Specifically, the agent is  given a detailed instruction, like \textit{``Head a bit ahead and towards the double doors on the left towards the kitchen. Stop upon reaching the counter.''}
At each step, then, the agent observes the panorama view of its surrounding environment and makes a decision for the direction to move in the next step, until it reaches the desired goal position. % point.

Recently, many methods~\cite{chattopadhyay2021robustnav,ke2019tactical,lin2021adversarial,nguyen2019vision,ma2019self,fried2018speaker,liu2021vision,hong2020language,tan2019learning,zhu2020vision,wang2019reinforced,zhao2021evaluation} have been proposed for the VLN task.
Most of the literature adopts the encoder-decoder framework to encode the instruction and visual observations, and then decode the action sequence.
Recent VLN studies \cite{qi2021know,li2019robust,majumdar2020improving,hao2020towards,guhur2021airbert} have shown great performance by directly modeling cross-modal vision-language modelling  with Transformer.
%However, there are two core challenges  worth exploring.
Different from other vision and language tasks, e.g. VQA and image captioning that learn relationships between each individual image and its corresponding text, VLN aims to learn the joint representation between each instruction and a series of observations by interacting with the environment.
Thus, taking the temporal context into account is the key to ground the instruction onto the observations, figuring out what has been completed, what is next, and where to go.
A straightforward way is to directly encode all the past observations~\cite{pashevich2021episodic}, which however misses record cross-modal history and also increases the training cost as the path grows.
% keep agent be aware of progress.
Further, \cite{hong2021vln} employs the recurrent hidden state to inject temporal information into Transformer and \cite{qi2021know,hao2020towards} use the encoder-decoder structure with an additional LSTM to encode the temporal context.
%Most of the existing VLN methods utilize a fixed-length latent vector to represent the temporal context.
%Meanwhile,it is different from the temporal video tasks, not all the temporal information is need.
%For example, \cite{hong2021vln} employs the recurrent hidden state to inject temporal information into Transformer and \cite{qi2021know,hao2020towards} use the encoder-decoder structure with an additional LSTM to encode the temporal context.
Nevertheless, a single hidden state vector is not expressive enough to encode the whole history of interactions with environment 
%which may lead to distance dependency 
in Transformer.
It is very challenging to align such hidden state at time $t$ with the corresponding sub-instruction for decision making.

To address this challenge, we propose a Multimodal Transformer with Variable-length Memory (MTVM) framework for VLN. 
Instead of using hidden states or an LSTM to encode temporal context, we find that it is simple and effective to directly reuse the cross-modal Transformer activations obtained in the previous steps.
Storing past activations in an explicit memory bank allows to explicitly model the cross-modal history.
%without the need to consider their distances in the path and 
Moreover, the Transformer architecture naturally accommodates variable-length memory token inputs. In this way, the agent is able to easily update the temporal context by adding the current output activation $m_t$, corresponding to the action at step $t$, into the memory bank, as shown in Figure~\ref{memoried transformer}.

Thanks to the explicit cross-modal memory bank, we further design a memory-aware consistency loss to boost the navigation performance.
The consistency loss aims to help cross-modal alignment by learning the relations between the previous activations and the language instruction.
Specifically, we randomly mask out some instruction words and force the model output distribution to be consistent with that of the original unmasked instruction.
%That is, even if some keywords are missing, the model can still localize the next step sub-instruction position in instruction according to the memory.
In this way, the model avoids overfitting to the language modality with the help of the explicit memory bank.

Our contributions can be summarized as follows:
\begin{enumerate}
\item We propose MTVM that allows the agent to capture temporal context without distance dependency by simply reusing the previous cross-model activations corresponding to the actions. 
%Note that a concurrent work HAMT~\cite{chen2021history}~\footnote{The paper has just been released in late Oct. 2021.} also proposes to model the history information explicitly. However, their method is fairly complex, requiring a hierarchical vision Transformer to encode intra-panorama and inter-panorama visual information for temporal context, whose end-to-end training is extremely costly and time-consuming.
\item We design a memory-aware consistency loss to learn strong relations between instruction and temporal context to further boost the navigation performance.
\item We conduct extensive experiments on R2R and CVDN datasets,  improving Success Rate by 2\% on R2R  and reducing Goal Progress by 1.5m on CVDN  compared to strong baselines.
%\jf{Pls double check your claim here.}.
%The results demonstrate the superiority of the proposed Memorized Transformer model.
\end{enumerate}

\section{Related Work}
\textbf{Vision-and-Language Navigation.} 
VLN~\cite{anderson2018vision} is a task that requires an agent to follow a nature-language instruction to navigate in a photo-realistic environment to a goal location.
In this process, the given instruction describes the trajectory in detail and the embodied agent needs to move through the scene with first person views as observations.
Following~\cite{anderson2018vision}, several navigation tasks~\cite{chen2019touchdown,thomason2020vision,nguyen2019help,nguyen2019vision,qi2020reverie} have been further proposed for interactions with surrounding environments. In particular, 
%if space limited
different from \cite{anderson2018vision} collecting data from an indoor environment, \cite{chen2019touchdown} extends the navigation environment to real-life visual urban streets. \cite{thomason2020vision} introduces navigating according to several question-answering pairs in a dialog history. 
\cite{kim2021ndh} further extends the dialog navigating task by taking the full dialogue and the whole navigation path as one instance.
\cite{nguyen2019vision} and \cite{nguyen2019help} consider object-finding tasks~\cite{ren2015faster,zhao2019cycleemotiongan,lin2020multi,lin2021domain} by requesting and interpreting simulated human assistants. \cite{qi2020reverie} requires the agent to navigate to an appropriate location and identify the target object.
\cite{ku2020room} proposes a multilingual datasets for VLN, which including more visual entities and avoiding language bias.

As a practical task in real-world applications, VLN has made incredible progress in recent years.
\cite{lin2021adversarial} uses adversarial attacking to capture key information from long instructions for a robust navigation.
%\cite{ma2019self} ensures grounding the instruction correctly with a progress monitor 
The progress monitor in \cite{ma2019self} aims to estimate the navigation progress explicitly as a multi-task learning, supervised by the normalized distance to the goal.
RCM~\cite{wang2019reinforced} enforces cross-modal grounding both locally and globally via a matching critic providing rewards for reinforcement learning.

In vision-and-language navigation setting, it is difficult to collect enough annotated data due to the large navigation space.
\cite{fried2018speaker} synthesizes new instructions where the speaker model helps the agent by additional route-instruction pairs to expand the limited training data.
To make further advances, \cite{zhao2021evaluation} proposes an instruction-trajectory compatibility model to improve the instruction evaluation.
\cite{tan2019learning} proposes an environmental dropout method based on the view consistency to mimic novel and diverse environments.
From a different perspective, REM~\cite{liu2021vision} reconnect the seen scenes to generate augmented data via mixing up environments.
To further understand the relations between the instructions and scenes, \cite{hong2020language} and \cite{qi2021know} take the objects in scenes and the corresponding words in instructions as the minimal units of encoding.
AuxRN~\cite{zhu2020vision} introduces additional training signals including explaining actions, predicting next orientation, etc.\ , to help acquire semantic knowledge.
In contrast, our method focuses on modeling the temporal context to help the alignment between language and observations.

\noindent
\textbf{Multi-Modal Transformers.}
The Transformer~\cite{vaswani2017attention} architecture has shown great effectiveness in vision and language tasks~\cite{tan2019lxmert,lu2019vilbert,chen2020uniter,li2020unicoder,huang2020pixel,zhou2020unified,gan2020large,referformer}. 
Most of the vision-and-language tasks focus on the joint embedding learning with individual pairs of an image and its corresponding language, such as VQA, image captioning, and text-to-image retrieval.
Different from these tasks, VLN is a Markov Decision Process, which learns the joint representation between the instruction and a series of observations along the corresponding trajectory.
Inspired by the success of BERT~\cite{devlin2018bert}, PRESS~\cite{li2019robust} first introduces a large-scale pre-trained language model to VLN for text representations.
As cross-modal joint learning is the key for VLN task, VLN-BERT~\cite{majumdar2020improving} and PREVALENT~\cite{hao2020towards} develop Transformer-based model in a self-supervised manner on image-text pairs from the web and image-text-action triplets from R2R dataset~\cite{anderson2018vision}, respectively.
\cite{hong2021vln} and \cite{qi2021know} adapt pre-trained V\&L BERT to VLN task by leveraging the hidden state representations with the learned linear projection or LSTM. Recently, HAMT~\cite{chen2021history} and Episodic Transformer~\cite{pashevich2021episodic} also propose to model the history information explicitly by directly encoding all past observations and the actions.
%, which however is fairly complex.
Our key insight is: \textit{only explicitly modelling the history observations is not good enough; instead, explicitly modelling the history interactions between observations and the instruction is more critical since it helps figure out the progress of the navigation trajectory.}
%Besides, we further design a memory-aware consistency loss to help the alignment between history information and language instruction.
%For example, HAMT requiring a hierarchical vision Transformer to encode intra-panorama and inter-panorama visual information for temporal context, whose end-to-end training is extremely costly and time-consuming.
%In contrast, our MTVM simply copies the past activations into a memory bank as the history information and we further design a memory-aware consistency loss to help the alignment between history information and language instruction.

\section{Methods}
\begin{figure*}[ht]
\centering
\includegraphics[width=1.0\linewidth]{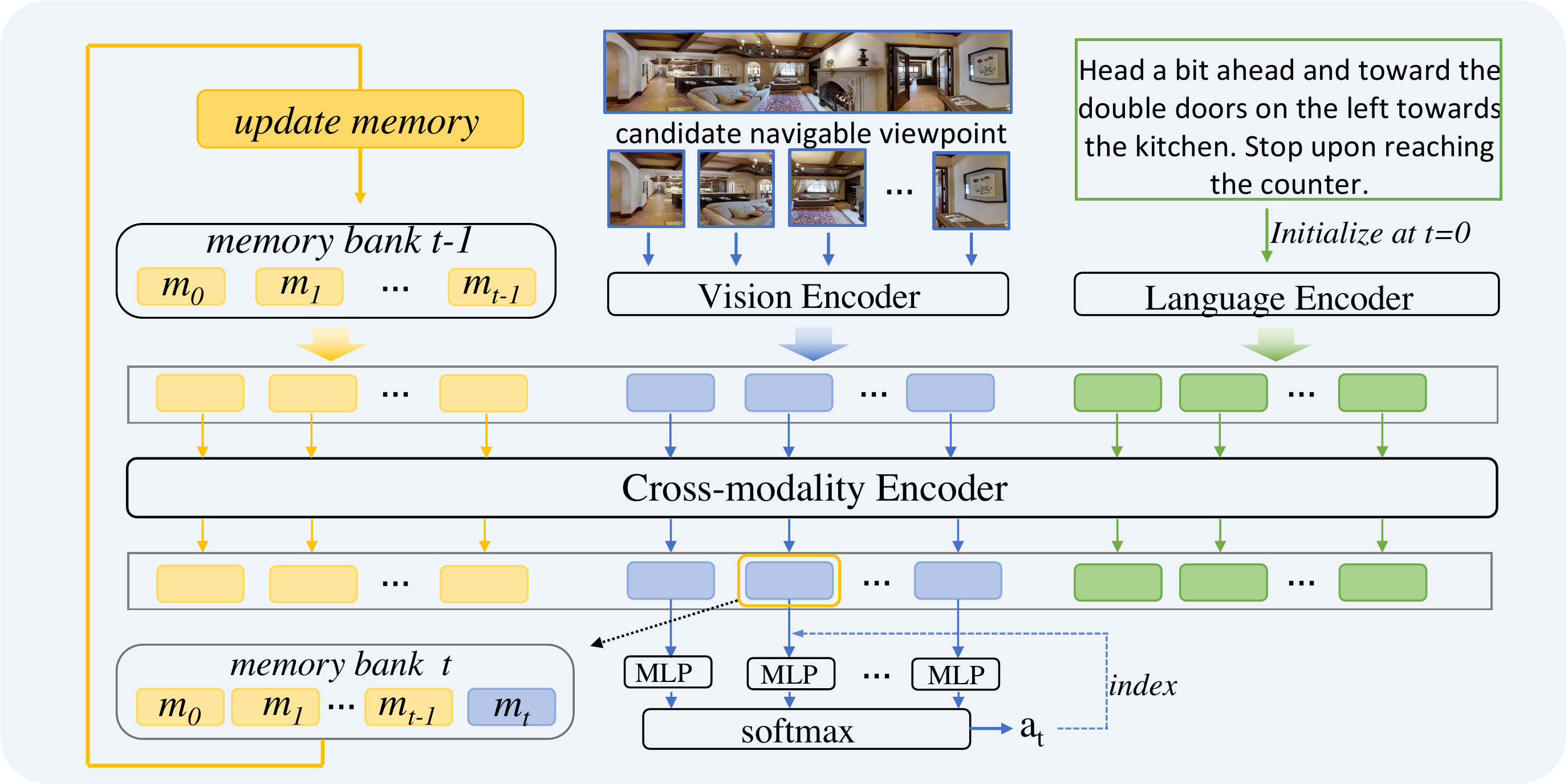}
\caption{The general framework of our proposed MTVM framework. At each step, we concatenate temporal context in the memory bank, together with visual features and language features as input. After making decision, we update the memory bank by storing the output activation that corresponding to the action.}
%Our method allows to model the cross-modal temporal context explicitly.} 
\label{framework_1}
\end{figure*}
\label{sec:methods}
\subsection{Overview}
Formally, at the beginning of each episode, the agent is given a nature language instruction $x = \left \langle x_1, x_2, \ldots, x_L \right \rangle$, where $L$ is the length of the instruction and $x_i$ denotes a word.
VLN task requires the agent to follow the instruction to navigate from a start position to the goal location.
At each step $t$, the agent is able to observe the surrounding environment in a panoramic view $o_t = \left \langle o_t^1, o_t^2, \ldots, o_t^{36} \right \rangle$ comprised by 36 single view images.
Figure \ref{framework_1} gives an overview of our proposed Multimodal Transformer with Variable-length Memory  (MTVM).
At each step, our MTVM directly interacts with visual information, language information, and history information to make the action decision.
%we feed vision feature, language feature and history activations into cross-modality encoder to interact with each other.
After that, we update the memory bank by reusing the activation of the Transformer output according to the action decision. Moreover, 
a consistency loss is introduced to measure the distance between the output distributions of the full instruction and a randomly masked instruction to help the cross-modal alignment. Note that the instruction masking  is only used in training but not in inference.
%During inference, we only use the full language path and output the agent's real moving trajectory.

\subsection{Memory-based Multimodal Transformer}
As VLN is a Markov decision process~\cite{anderson2018vision}, an embodied agent needs to pay attention to the temporal context information during its navigation.
The general Transformer is not enough to model the instruction and the observations due to the lack of the temporal context.
At each navigation step, an agent needs to ground an instruction to which part has finished and which part is the next.

MTVM learns the cross-modal alignment to encourage matching the completed part of the instructions with the past trajectory.
%enables the direct connects between previous activations and language embedding to make the completed part matched with the past trajectory.
Our memory bank enables the agent to be aware of the navigation process by directly interacting with the previous actions so that it can ground the sub-instructions as guidance.
In this way, it becomes easier for the agent to locate the sub-instruction to gain useful information to select the candidate direction from the current-step observation.
We construct our model following the vision and language pretrained work~\cite{tan2019lxmert,hao2020towards}, which consists of a language encoder, a vision encoder and a cross-modality encoder.

\begin{figure*}[t]
\centering
\includegraphics[width=1.0\linewidth]{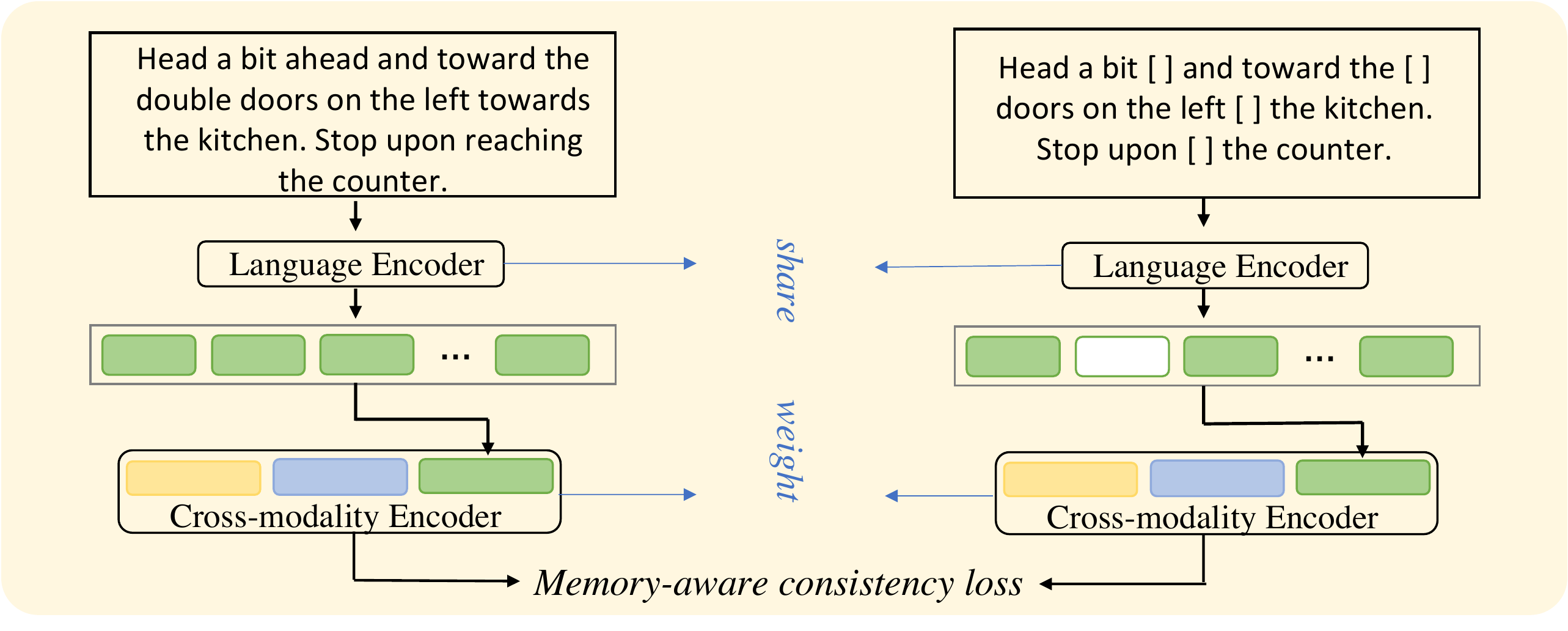}
\caption{The proposed memory-aware consistency loss. During training, we randomly mask out some words to help the alignment between language and temporal context, avoiding model overfitting to the language modality.} 
\label{framework}
\end{figure*}

\noindent \textbf{Language Encoder.}
The language encoder is a standard multi-layer transformer with self-attention. 
At the beginning of an episode, we feed the instruction to the language encoder $\mathcal{S}$ to get the language representation $X =\mathcal{S}(x)$.

\noindent \textbf{Vision Encoder.}
The vision encoder is a convolution network to encode each single view image $o_t^i$ to a 2048-dimensional visual feature $v_t^i$.
A 128-dimensional directional feature $d_t^i$ by repeating the trigonometric function representation~\cite{fried2018speaker} is concatenated with the visual feature $v_t^i$ to represent the orientation of each single view $V_t^i = [v_t^i; d_t^i]$.
%\jf{Better give a reference here}
%$[\sin{\psi_t^i}; \cos{\psi_t^i}; \sin{\theta_t^i}; \cos{\theta_t^i}]$
%Empirically, we only use the visual feature $V_t = [v_t; d_t]$
For each step, we have $V_t = \{V_t^1, V_t^2, \ldots, V_t^K\}$ as the visual representation, where $K$ is the number of candidate directions.
%\jf{Symbols V do not match with those in Sec. 3.1.}.

\noindent \textbf{Cross-modality Encoder.}
In order to learn cross-modality representations, the cross-modality encoder $\mathcal{C}$ is composed of self-attention layers and cross-attention layers, where cross-attention layers treat one modality as query and the other as key and value to exchange the information and align the entities between the two modalities.
%Besides position embedding as normal, we add the history mark and current mark as type embedding to the history token and vision token.
In particular, we feed language representation $X$, vision representation $V_t$, and previous activations $M_t$ to the cross-modality encoder $\mathcal{C}$ as
\begin{equation}
\widehat{X}, \widehat{M_t}, \widehat{V_t} = \mathcal{C}(X, [M_t; V_t]),
\label{equ1}
\end{equation}
where $[;]$ denotes concatenation.
Then, the action prediction head takes the output $\widehat{V_t}$ to make the action decision for this step: $a_t = MLP(\widehat{V_t})$.

At the end of each step, we update the memory bank by reusing the output activations $\widehat{V_t}^k$ according to the current agent action decision as
\begin{equation}
M_t \leftarrow (M_{t-1}, \left[\widehat{V_t}^k;d^k_t\right])
\label{equ2}
\end{equation}
where $k$ is the index of the selected vision output and $d_t^k$ is the corresponding directional feature of $t$ step action.
%\jf{In Sec 3.1 you use $\theta$ and $\phi$, but here you use $d$. Be consistent!}.

\subsection{Memory-aware Consistency Loss}
%As discussed above, the key challenge in VLN is embodied agents need to be aware of the progress of navigating trajectory in order to ground the sub-instruction as guidance.
As aforementioned, the key challenge in VLN is that the embodied agent needs to be aware of the progress of the navigating trajectory by learning the cross-modal representation.
%the alignment between temporal context and language embedding.
However, the existing studies~\cite{hu2019you,anderson2018vision} show that the agent tends to overfit the instructions, which could be due to large variations in the visual modality. 
%as the visual groundings learned may be quite specific to the training environments, the same foundation can be seen in \cite{hu2019you,anderson2018vision}.
%Although considering each past step in memory bank has made instruction-and-trajectory alignment easier, the model may still only capture language instructions, such as route words (turning left or moving forward), the same foundation can be seen in \cite{hu2019you}.
In order to avoid the model from overfitting a single modality, we design a memory-aware consistency loss. % to facilitate the model to better generalize to the unseen environments.
By randomly dropping some words in the instruction, we force the model to learn strong representations among language, vision, and temporal context from the cross-modality encoder.
%close the distance between outputs distributions to enhance the cross-modal alignment.

Specifically, given an instruction $x$, we random drop some words with a fixed probability and obtain
\begin{equation}
x' = RandomDrop(x).
\label{equ3}
\end{equation}
Both $x$ and $x'$ are then encoded by language encoder $\mathcal{S}$ to produce the instruction representations $X$ and $X'$, respectively.
Same as the instruction feature $X$, $X'$ is also fed through the cross-modality encoder $\mathcal{C}$ with the same history and vision representations as Eq.~\eqref{equ1}:
\begin{equation}
\widehat{X'}, \widehat{M'_t}, \widehat{V'_t} = \mathcal{C}(X', [M_t; V_t]),
\label{equ4}
\end{equation}
%As discussed above, the masked instruction force the model to learn the strong representation.
Although some words are discarded, we expect the similarities between the instruction features $X$ and $X'$ and their corresponding outputs are preserved. 
%the model should predict the similar alignment for two modality tokens. Our model tries to measure the similarities of cross-modal alignment in a representation space.
%Concretely, we minimize the bidirectional Kullback-Leibler (KL) divergence between the outputs of the full instruction and the randomly dropped instruction from the language encoder and the cross-modality encoder by generating the probability vectors with a Softmax layer.
Concretely, we generate the probability vectors for the outputs of the language encoder and cross-modality encoder respectively with the Softmax layer.
By minimizing the bidirectional Kullback-Leibler (KL) divergence between the outputs of the full instruction and the randomly dropped instruction, the consistency loss is defined as
\begin{equation}
\begin{split}
\mathcal{L}_{consis} = \lambda_{s} (\mathcal{D}_{KL}(X\|X') + \mathcal{D}_{KL}(X'\|X)) &\\
+ \lambda_{m}(\mathcal{D}_{KL}(\widehat{X'}, \widehat{M'_t}, \widehat{V'_t}\|\widehat{X}, \widehat{M_t}, \widehat{V_t}) + &\\
\mathcal{D}_{KL}(\widehat{X}, \widehat{M_t}, \widehat{V_t}\|\widehat{X'}, \widehat{M'_t}, \widehat{V'_t}))&,
\end{split}
\label{equ5}
\end{equation}
where $\lambda_{s}$ and $\lambda_{m}$ are the weights to balance the distance losses.
%in Eq.~\eqref{equ5}
The first term aims to prevent the agent from overfitting the special words (such as route words), while the second term aims to avoid overfitting the language modality.

\subsection{Training}
Following the existing VLN works, we apply the mixture of Imitation Learning (IL) and Reinforcement Learning (RL)  strategies~\cite{wang2019reinforced,tan2019learning}.
In IL, the agent learns to follow the teacher action $a_t^*$ of the ground-truth path at each step $t$ by minimizing the negative log probability loss function.
In RL, the agent learns from rewards by using A2C algorithm~\cite{mnih2016asynchronous}, where sampling the action $a_t^s$ from the agent's  action distribution $a_t$, the agent will get rewards if successfully arriving at the target within 3m ($t=T$) or reducing the distance to the target after taking the action ($t<T$).
Besides, we consider the similarity of the agent path and the ground-truth path as a reward to encourage the agent follow the instruction to move closer to the target.
The overall loss function can be written as:
\begin{equation}
\begin{split}
\mathcal{L} &=\lambda_{l}\mathcal{L}_{IL} + \mathcal{L}_{RL}+ \mathcal{L}_{consis}\\
&=\lambda_{l}\sum_{t=0}^{T-1} -a_t^*log(a_t) + \sum_{t=0}^{T-1}-a_t^slog(a_t)A_t + \mathcal{L}_{consis}\\
\end{split}
\label{equ6}
\end{equation}
where $\lambda_{l}$ is a trade-off weight for IL loss, $T$ is the length of the navigation path, and $A_t$ is the advantage calculated by A2C algorithm~\cite{mnih2016asynchronous}.
We alternately train the agent with IL and RL strategies while applying the consistency loss in both.

\section{Experiments}
%In this section, we introduce the datesets we use to evaluate our model, evaluation metrics, and implementation details.
\subsection{Setup}
\textbf{Datasets:}
We evaluate MTVM on the Room-to-Room dataset (R2R)~\cite{anderson2018vision} and Cooperative Vision-and-Dialog Navigation dataset (CVDN)~\cite{thomason2020vision} in 3D environments based on Matterport3D Simulator~\cite{chang2017matterport3d}.
The simulated environments include 90 different housing scenes.
R2R dataset provides fully specified instructions describing the steps necessary to reach the goal, while CVDN dataset provides an ambiguous and underspecified goal location and human-human dialogs to guide the agent.
R2R splits the dataset into the training set consisting of 61 environments with 14,025 instructions, the seen validation set consisting of the same 61 environments with 1,020 instructions, and the unseen validation consisting of another 11 environments with 2,349 instructions, while the test consists of the remaining 18 environments with 4,173 instructions.
CVDN contains 4742 training, 382 seen validation, 907 unseen validation, and 1384 unseen test instances.

\textbf{Evaluation Metrics:}
For R2R, we use its three standard metrics: Navigation Error (NE) defined as the distance (in meters) from the stop viewpoint to the goal position, Success Rate (SR), and Success rate weighted by Path Length (SPL), where SPL is regarded as the primary metric. 
For CVDN, following~\cite{thomason2020vision}, we evaluate the performance on the navigation from dialog history (NDH) task by Goal Progress, which measures how much reduction in meters the agent makes towards the goal.
There are three settings depending on the supervised strategy. 
\textit{Oracle} indicates the agent regarding the shortest path as ground truth and
\textit{Navigator} indicates learning from the navigator path (maybe not be the optimal navigation).
\textit{Mixed} supervision means to learn from the navigator path if it reaches the goal point; otherwise learn from the shortest path.

\textbf{Implementation Details:}
To leverage vision and language pre-trained models, we initialize the language encoder and the cross-modality encoder by a pre-train VLN model PREVALENT~\cite{hao2020towards}.
Following PREVALENT~\cite{hao2020towards} and VLN$\circlearrowright$BERT~\cite{hong2021vln}, we train the agent on the original training data and the augmented data provided by \cite{hao2020towards}.
The vision encoder is a fixed ResNet-152~\cite{he2016deep} pre-trained on Place365~\cite{zhou2017places} provided by R2R dataset.
The experiments are conducted on 3 V100 GPUs. 
We train the model 10,000 iterations and adopt the early stopping strategy when the model achieves the best performance on the evaluation metric.
The learning rate is fixed to $5\mathrm{e}{-6}$ with an AdamW optimiser~\cite{loshchilov2017decoupled}.
The parameters $\lambda_s$ and $\lambda_m$ are respectively set to 0.6 and 0.2 %to balance the loss of language encoder and cross-modality encoder 
and $\lambda_{IL}$ is set to 0.2. % to weight IL learning loss.
We find different levels of dropping words are all helpful, and we fix the word dropping probability to 0.5.

\setlength{\tabcolsep}{4pt}
\begin{table}[tpb]
\centering
\caption{Comparisons of the VLN performance on R2R dataset in a single-run setting. The best results are in bold font. The set of methods at the bottom are Transformer based solutions, whose model parameters are initialized by the pre-trained vision-and-language BERT. The set of methods in the middle are non-Transformer based solutions.}
%Note that for the very recent concurrent method HAMT~\cite{chen2021history}, we report its results with Resnet-152 as the vision encoder for fair comparison}
%NE: SR: SPL:
\begin{tabular}{@{}l|ccc|ccc|ccc@{}}
\hline
\hline
\multirow{2}*{Methods} & \multicolumn{3}{|c|}{Validation Seen} & \multicolumn{3}{|c|}{Validation Unseen} & \multicolumn{3}{|c}{Test} \\
\cline{2-10}
~&NE$\downarrow$&SR$\uparrow$&SPL$\uparrow$ &NE$\downarrow$&SR$\uparrow$&SPL$\uparrow$ &NE$\downarrow$&SR$\uparrow$&SPL$\uparrow$ \\
%~&NE(m)$\downarrow$&SR(\%)$\uparrow$&SPL(\%)$\uparrow$ &NE(m)$\downarrow$&SR(\%)$\uparrow$&SPL(\%)$\uparrow$ &NE(m)$\downarrow$&SR(\%)$\uparrow$&SPL(\%)$\uparrow$ \\
\hline
\hline
Random&9.45&16&-&9.23&16&-&9.79&13&12\\
Human&-&-&-&-&-&-&1.61&86&76\\
\hline
Speaker-Follower~\cite{fried2018speaker} &3.36&66&-&6.62&35&-&6.62&35&28\\
Self-monitoring~\cite{ma2019self}&3.22&67&58&5.52&45&32&5.67&48&35\\
RCM~\cite{wang2019reinforced} &3.53&67&-&6.09&43&-&6.12&43&38\\
FAST-Short~\cite{ke2019tactical}&-&-&-&4.97&56&43&5.14&54&41\\
EnvDrop\cite{tan2019learning} &3.99&62&59&5.22&52&48&5.23&51&47\\
DR-Attacker~\cite{lin2021adversarial} &3.52&70&67&4.99&53&48&5.53&52&49\\
AuxRN~\cite{zhu2020vision} &3.33&70&67&5.28&55&50&5.15&55&51\\
RelGraph~\cite{hong2020language} &3.47&67&65&4.73&57&53&4.75&55&52\\
\hline
PRESS~\cite{li2019robust}&4.39&58&55&5.28&49&45&5.49&49&45\\
PREVALENT~\cite{hao2020towards}&3.67&69&65&4.71&58&53&5.30&54&51\\
ORIST~\cite{qi2021know} &-&-&-&4.72&57&51&5.10&57&52\\
VLN$\circlearrowright$BERT~\cite{hong2021vln}&2.90&72&68&3.93&63&57&4.09&63&57\\
%HAMT$^{*}$~\cite{chen2021history}&-&69&65&-&64&58&-&-&-\\
%Mixup*~\cite{liu2021vision}&\textbf{2.48}&\textbf{75}&\textbf{72}&3.89&64&58&\textbf{3.87}&\textbf{65}&\textbf{59}\\
%Ours&2.71&74&70&\textbf{3.73}&\textbf{65}&\textbf{59}&3.88&\textbf{65}&\textbf{60}\\
Ours&\textbf{2.67}&\textbf{74}&\textbf{69}&\textbf{3.73}&\textbf{66}&\textbf{59}&\textbf{3.85}&\textbf{65}&\textbf{59}\\
%HAMT (ViT) &2.51 &76 &72 & 2.29 & 66 & 61 & 3.93 & 65 & 60\\
\hline 
\hline 
\end{tabular}
\label{table1}
\end{table}
\setlength{\tabcolsep}{1.4pt}

\subsection{Comparisons with SoTA}
Table \ref{table1} shows the performance comparisons of different VLN methods on R2R dataset in a single-run setting. 
It can be seen that our model performs the best on all the metrics under both unseen validation and test sets, suggesting the good generalizing ability.
%We initialize the model weight from PREVALENT~\cite{hao2020towards}, which is a pre-train model for the VLN problem, and our method achieves 11\% and 9\% improvement on test set success rate and SPL. 
Compared with other transformer-based methods including PRESS~\cite{li2019robust}, ORIST~\cite{qi2021know} and VLN$\circlearrowright$BERT~\cite{hong2021vln} which also initialize their models using the pre-trained ones~\cite{hao2020towards,chen2020uniter}, our method is at least 2\% higher in terms of SPL or SR under both test and validation unseen scenarios.
In addition, the lowest navigation error achieved by our model indicates that we can make the agent move closer to the target.
%Note that for the very recent concurrent work HAMT~\cite{chen2021history}, we report its results with Resnet-152 as the vision encoder for fair comparison. When changing Resnet-152 to ViT~\cite{dosovitskiy2020image}, HAMT reports better performance. However, its end-to-end training requires 20 NVIDIA V100 GPUs for $\sim$ 20 hours, which is much higher than ours (3 V100 for 1 day). Another recent work Mixup~\cite{liu2021vision} extends 61 scenes in the training set to 116 cross-connected scenes with data augmentation and thus uses more data than other methods. 
%HAMT~\cite{chen2021history} replace the visual feature from Resnet-152 to ViT~\cite{dosovitskiy2020image} to get a better performance.
%, and hence is not directly equivalent.
%For fair comparison, we report HAMT$^*$ results which use Resnet-152 as vision encoder.
% We leave more comparisons in Appendix. %\jf{Make sure you will do it if you write something here.}
%Our method achieves state-of-the-art results on the unseen scenario even take Mixup into account.

\begin{table}[tpb]
\centering
\caption{Comparisons with state-of-the-art methods in terms of Goal Progress (m) on the navigation from dialog history (NDH) task on CVDN dataset~\cite{thomason2020vision}. `Ora', `Nav' and `mix' denote the three settings, `Oracle', `Navigator' and `Mixed', respectively.}
\begin{tabular}{@{}l|ccc|ccc@{}}
\hline
\hline
\multirow{2}*{Methods}  & \multicolumn{3}{c}{Validation Unseen} & \multicolumn{3}{|c}{Test} \\
\cline{2-7}
~&Ora&Nav&Mix &Ora&Nav&Mix\\
%~&Oracle&Navigator&Mixed &Oracle&Navigator&Mixed \\
\hline
\hline
Random&1.09&1.09&1.09&0.83&0.83&0.83\\
Shortest Path&8.36&7.99&9.58&8.06&8.48&9.76\\
\hline
Seq-to-seq~\cite{thomason2020vision} &1.23&1.98&2.10&1.25&2.11&2.35\\
PREVALENT~\cite{hao2020towards} &2.58&2.99&3.15&1.67&2.39&2.44\\
CMN~\cite{zhu2020vision1}&2.68&2.28&2.97&2.69&2.26&2.95\\
ORIST~\cite{qi2021know}&3.30&3.29&3.55&2.78&3.17&3.15\\
SCoA~\cite{zhu2021self} & 1.94 & 2.91 & 2.85 & 2.49 & 3.37 & 3.31 \\
DR-Attacker~\cite{lin2021adversarial}&3.27&4.00&4.18&2.77&2.95&3.26\\
Ours&\textbf{4.57} &\textbf{4.80} &\textbf{5.15} &\textbf{4.23} &\textbf{4.46} &\textbf{4.82}\\
\hline 
\hline 
\end{tabular}
\label{table2}
\end{table}

Table~\ref{table2} shows the performance comparisons in terms of Goal Progress on CVDN dataset under the three different settings.
Again, our method achieves the best performance with significant gains on both unseen validation and test sets, demonstrating the effectiveness of handling a variety of language instructions.
Note that the Shortest Path Agent takes the shortest path to the supervision goal at inference, which represents the upper bound navigation performance for an agent.
%Compared to the previous best result, our model has improved significantly on both validation unseen and test sets.
%On the validation unseen set, the Goal Progress reduce 1.20m, 0.80m, and 0.97m respectively.

\subsection{Ablation studies}
\begin{figure*}[t]
\centering
\includegraphics[width=1.0\linewidth]{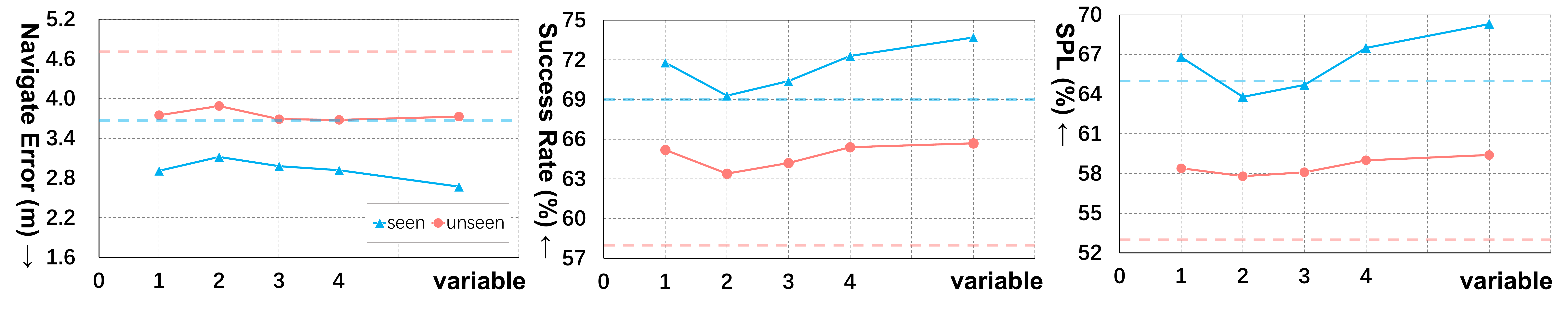}
\caption{Impacts of the memory bank size on seen and unseen validation sets of R2R dataset in terms of NE, SR and SPL. Solid lines are our results with different memory bank sizes, and dashed lines are the results of PREVALENT~\cite{hao2020towards} from which our model is initialized.} 
\label{mem size}
\end{figure*}

\textbf{Memory bank size.} Recall that our method stores the activations at each step as history information in a memory bank. Here, we evaluate the model performance with different memory bank sizes. When the memory bank size is $n$, we only record the last $n$ step activations; when the size is variable, it means we record every step. Note that the paths in R2R dataset are all around four to six steps.
%We record each step activations in our work to align the language instruction with history information. Noted paths in R2R dataset is all four to six steps. To understand how this strategy affects model performance, we evaluate memory bank with fixed size \{1 $\sim$ 4\}. When memory bank size is $n$, we only record the last $n$ step activations, and variable-length means recording every step.
The results are shown in Figure~\ref{mem size}. In general, a larger memory size helps, and the variable-length memory gives the best performance, suggesting the importance of explicitly storing the history information.
In addition, we also show the performance of PREVALENT as our baseline (dashed lines) since our model is initialized from it. It can be seen that our model under most of the fixed-length memory banks outperforms the baseline.
%\jf{In the caption of Fig. 3, you say the dashed lines are the results of VLNBERT. Inconsistent!}
%Most of the fix length memory bank improves performance demonstrating the effectiveness of our proposed method.

\noindent
\textbf{Comparison of history encoding methods.}
We next evaluate the advantage of proposed variable-length memory bank to other baselines, including visual-only \cite{pashevich2021episodic,chen2021history} and cross-modal interaction~\cite{hong2021vln} as history encoding methods.
%, as measured by Success Rate(SR) and SPL in R2R validation unseen setting.
\cite{pashevich2021episodic} encodes oriented observations (one view of full observations) and the actions as the history information.
\cite{chen2021history} proposes a hierarchical observations and actions encoding method which is able to learning intra-panorama and inter-panorama visual information for temporal context.
The experiments are under the R2R validation seen and unseen setting and measured by Success Rate (SR) and Success rate weighted by Path Length (SPL).
As \cite{pashevich2021episodic} was proposed for the ALFRED benchmark which is for household action learning from instructions and egocentric vision, we reproduce it by simply replacing our history encoding with the corresponding methods.

As shown in Table\ref{table3}, we have the following observations from the results:
(1) The visual-only method \cite{pashevich2021episodic} that only encodes the past oriented views and actions obtains the worst performance.
This is obvious, because only recording the oriented views may ignore significant information in the trajectory.
For instance, ``Go straight passing the fridge", ``fridge" might not be in the oriented visual observations, which is essential for the agent to record history.
(2) Compared with visual-only methods, the cross-modal history encoding methods achieves better performance in most settings, which demonstrates the effectiveness of considering the multimodal interactions as history for VLN.
Only modelling history observations provides the visual information in temporal context, but is insufficient to record vision and language navigation progress.
(3) Our MTVM achieves the highest SR and SPL among all history encoding methods, because of the proposed variable length memory and the memory consistency loss.
Compared to the typically used recurrent state, we found that cross-modal history can be better captured by simply reusing the previous cross-model activations corresponding to the actions, which is simple but effective and non-trivial. 
Note that for HAMT~\cite{chen2021history}, we report its results with Resnet-152 as the vision encoder for fair comparison. 
%When changing Resnet-152 to ViT~\cite{dosovitskiy2020image}, HAMT reports better performance. 
%However, its end-to-end training requires 20 NVIDIA V100 GPUs for $\sim$ 20 hours, which is much higher than ours (3 V100 for 1 day).

\begin{table}%[htpb]
\centering
\setlength{\tabcolsep}{1.5mm}
\caption{Comparison of different history encoding methods in R2R setting. ``Visual-only" indicates methods that encoding past observations and actions as history. ``Cross-modal" indicates methods considering cross-modal interactions and actions as history.}
\begin{tabular}{|l|l|cc|cc|}
\hline
\multicolumn{2}{|l}{\multirow{2}{*}{History Encoding Methods}}& \multicolumn{2}{|c}{Val Seen} & \multicolumn{2}{|c|}{Val Unseen} \\
\cline{3-6}
\multicolumn{2}{|c|}{} &SR$\uparrow$&SPL$\uparrow$ &SR$\uparrow$&SPL$\uparrow$\\
\hline
\multirow{2}*{Visual-only} & E.T.~\cite{pashevich2021episodic} \textit{Oriented observations} &68.1 & 63.6& 59.0 & 54.5 \\
%~&Panoramic observations & 59.4 & 54.3 \\
~&HAMT~\cite{chen2021history} \textit{Hierarchical observations}& 69.3 & 64.8 & 63.5 & 57.5 \\
\hline
\multirow{2}*{Cross-modal} & VLN$\circlearrowright$BERT~\cite{hong2021vln} \textit{Recurrent state} & 72 & 68 & 63 & 57 \\
~&\textbf{Ours} & \textbf{73.7} & \textbf{69.3} & \textbf{65.7} & \textbf{59.4} \\
\hline
\end{tabular}
\label{table3}
\end{table}

\begin{table}%[htpb]
\centering
\setlength{\tabcolsep}{2mm}
\caption{Impacts of our proposed memory-aware consistency loss and random word dropping. ``word dropping'' refers to our model without using the consistency loss but with random word dropping in language instructions for data augmentation.}
\begin{tabular}{ccc|cc}
\hline
\multicolumn{3}{c|}{Methods} & \multicolumn{2}{c}{Validation Unseen} \\
\hline
memory bank& consistency & word dropping & SR(\%)$\uparrow$ & SPL(\%)$\uparrow$\\
\hline
\checkmark & & & 64.0 & 58.6 \\
\checkmark & \checkmark & & 65.7$_{\uparrow 1.7}$ & 59.4$_{\uparrow 0.8}$\\
\checkmark & & \checkmark & 64.5$_{\uparrow 0.5}$ &57.8$_{\downarrow 0.8}$ \\ 
\hline
\end{tabular}
\label{table4}
\end{table}

\begin{table}[tpb]
\centering
\caption{Comparisons of training memory and computation cost on R2R dataset. We produce MTVM$\dagger^*$ with the same cross-attention strategy as VLN$\circlearrowright$BERT, where language is used as keys and values but not as queries. $\dagger$ indicates MTVM without the consistency loss. 
The best results are in bold and the second best results are underlined. }
\begin{tabular}{l|c|c|cc}
\multicolumn{3}{c}{}\\
\hline
\multirow{2}*{Methods} & \multirow{2}*{Params$^{\#}$} & \multirow{2}*{Memory} & \multicolumn{2}{c}{Validation Unseen} \\
\cline{4-5}
& & &SR(\%)$\uparrow$&SPL(\%)$\uparrow$\\
\hline
VLN$\circlearrowright$BERT &41.9M & 8.6GB & 63.3 & 57.5\\
MTVM$\dagger^*$ & 41.6M & 8.4GB & \underline{63.6} & \underline{58.2} \\
MTVM$\dagger$ & 68.4M & 17.9GB & \textbf{64.0} & \textbf{58.6} \\
\hline
\end{tabular}
\label{table5}
\end{table}

\iffalse
\begin{table}%[htpb]
\centering
\caption{Impacts of our proposed memory-aware consistency loss and random word dropping. ``MTVM w/o consistency + drop words'' refers to our model without using the consistency loss but with random word dropping in language instructions for data augmentation. }
\begin{tabular}{l|cc}
\multicolumn{3}{c}{}\\
\hline
\multirow{2}*{Methods} & \multicolumn{2}{c}{Validation Unseen} \\
\cline{2-3}
～&SR(\%)$\uparrow$&SPL(\%)$\uparrow$\\
\hline
%VLN$\circlearrowright$BERT~\cite{hong2021vln}&63.3&57.5\\
%\rowcolor{mygray} VLN$\circlearrowright$BERT~\cite{hong2021vln} + consistency&62.8$_{\downarrow 0.5}$&57.2$_{\downarrow 0.3}$\\
%Lxmert~\cite{tan2019lxmert}&44.9&40.6\\
%\rowcolor{mygray} Lxmert~\cite{tan2019lxmert} + consistency&46.5$_{\uparrow 1.6}$&40.9$_{\uparrow 0.3}$\\
MTVM w/o consistency & 64.0 & 58.6 \\
\rowcolor{mygray}
MTVM & 65.7$_{\uparrow 1.7}$ & 59.4$_{\uparrow 0.8}$ \\
\hline
MTVM w/o consistency + drop words &64.5   &57.8 \\ \hline
\end{tabular}
%, we add it to the VLN$\circlearrowright$BERT as well. The results demonstrate that the consistency loss can only learn the cross-modal joint representation with help of the explicit memory bank.}
\label{table4}
\end{table}
\fi

\noindent
\textbf{Impacts of consistency loss and random word dropping.}
Table~\ref{table4} compares the results with and without our proposed consistency loss. For our MTVM model, we can see that the consistency loss significantly improves the performance. 
The consistency loss is designed to encourage the model to pay more attention to our explicitly modelled history tokens.
Although some words are dropped during training, a lot of vision-language alignments have already been captured in the memory. 
It improves 1.7$\%$ and 0.8$\%$  on R2R validation unseen setting with SR and SPL metric, indicating that the agent with the memory consistency loss achieves better generalize ability.
%We also evaluate its impact to another method, VLN$\circlearrowright$BERT, in Table~\ref{table4}. 
%To further understand the consistency loss, we analyze the impact of adding it to other method in Table \ref{table4}.
%Specifically, we use the same word dropping strategy to VLN$\circlearrowright$BERT with the consistency loss.
%However, for VLN$\circlearrowright$BERT, adding the consistency loss performs slightly worse, suggesting hard to learn cross-modal representations without explicit memory bank.

%\textbf{Random Word dropping.} 
%For each instruction, we randomly drop some words to help model learning consistency representation. 
Note that our word-drop strategy for the consistency loss is similar to conventional random word dropping used for data augmentation. Thus, we make a comparison with direct word dropping for data augmentation (denoted as ``memory bank" + ``word dropping'') in Table~\ref{table4}, where 
we fix the word dropping rate to 0.5 in all methods. It can be seen that direct word dropping as data augmentation is not as effective as ours.

\begin{figure}[t]
\centering
\includegraphics[width=1.0\linewidth]{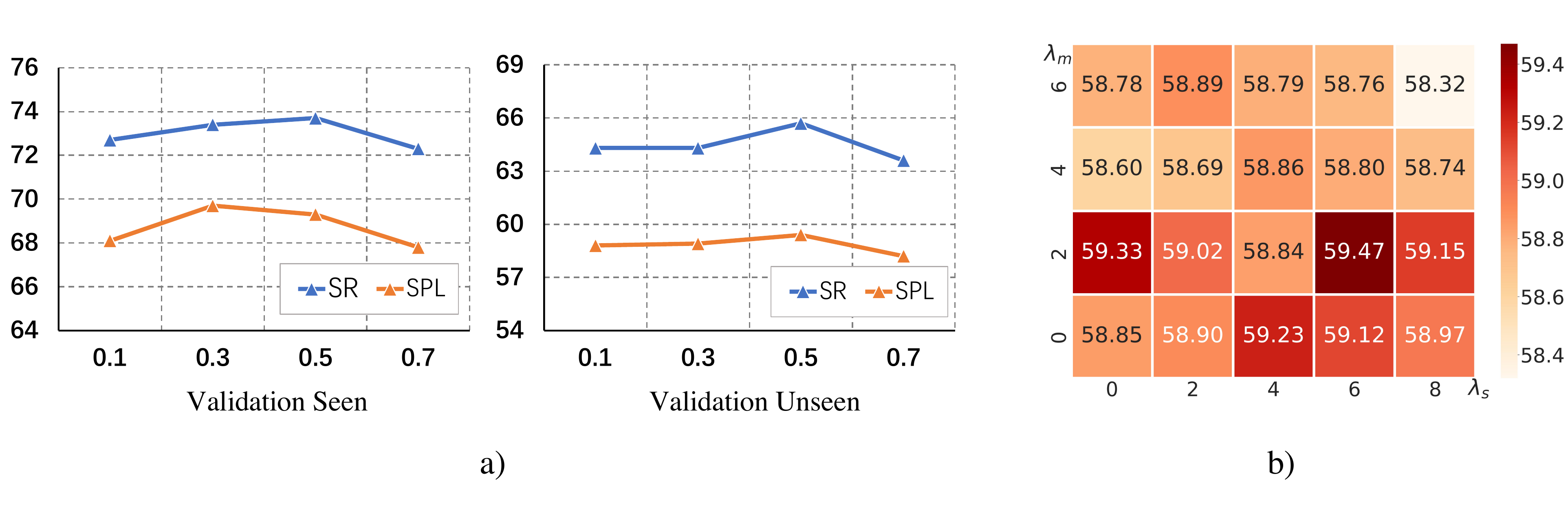}
\caption{a) Impact of different random word dropping rates on SR
and SPL on both seen and unseen validation sets of R2R dataset. b) Sensitivity examples of the hyper-parameters in Eq.~\eqref{equ5} to SPL metric on R2R unseen validation set. The darker the color, the better the performance.} %We chose the parameters $\lambda_{s}=6, \lambda_{m}=2$ with the best results.} 
\label{para}
\end{figure}

%\textbf{Random Drop Rates.} 
We further investigate the effect of different word dropping rates on SR and SPL in both seen and unseen validation sets of R2R dataset. 
Here we conduct experiments by varying the word dropping rate in $\{0.1, 0.3, 0.5, 0.7\}$.
As shown in Fig~\ref{para} a), we can see that a small dropping rate (e.g., 0.1) does not perform as good as a large one (e.g., 0.5),  %, which means slight drop words has little help in cross-modality alignment.
while a too large dropping rate (e.g. 0.7) also hurts the performance. Thus, the best choice is 0.5.
\iffalse
\begin{figure}[htpb]
\centering
\includegraphics[width=1.05\linewidth]{Figure/Fig4.pdf}
\caption{Impact of different random word dropping rates on SR and SPL on both seen and unseen validation sets of R2R dataset.} 
\label{drop rate}
\end{figure}
\fi

\noindent
\textbf{Hyper-parameter sensitivity.} 
We analyze the sensitivity of the hyper-parameters to SPL metric on R2R unseen validation set by using $\lambda_s$ and $\lambda_m$ in Eq.~\eqref{equ5} as examples.
The results are reported in Figure~\ref{para} b). 
%Among these different results, we can see that the SPL values achieve a stably improvement in the range of 2 $\sim$ 8.
%One interesting point is even one of the parameters is set to 0, it can still obtain a satisfying result which indicates consistency loss can learn strong representation from both language instruction and cross-modality sides.
From these results, we can see that SPL is not very sensitive to the variations of $\lambda_s$ and $\lambda_m$ in a range around 2 $\sim$ 8 and we find that it is a good choice to set $\lambda_{s}=6, \lambda_{m}=2$.

\noindent
\textbf{Memory and computation Cost.}
Following most of the cross-modal Transformer methods~\cite{tan2019lxmert,hao2020towards}, our MTVM facilitates vision-and-language interactions by bi-directional cross-attention sub-layers, where language is used as query attending to vision and vice versa. To compare with single-direction cross-modal Transformer method VLN$\circlearrowright$BERT~\cite{hong2021vln}, which only considers language tokens as keys and values but not as queries, we also develop a similar version, MTVM$\dagger^*$. 
%To reduce the memory consumption, we develop  MTVM$\dagger^*$ only considered the language token as keys and values but not queries, which is similar to VLN$\circlearrowright$BERT~\cite{hong2021vln}.
The comparison results of  VLN$\circlearrowright$BERT,  MTVM$\dagger^*$ and MTVM$\dagger$ in terms of Parameters and GPU Memory Cost are shown in Table \ref{table5}.
For a fair comparison with VLN$\circlearrowright$BERT, all the experiments are conducted on a single V100 GPU with batch size 16.
With the same cross-attention strategy, compared with VLN$\circlearrowright$BERT, our MTVM$\dagger^*$ archives better performance but with lower memory and computation cost. This is because VLN$\circlearrowright$BERT needs an additional small network to encode update its hidden states for temporal context while our MTVM$\dagger^*$ directly reuses the previous activations. This demonstrates the efficiency and effectiveness of our proposed memory bank based Transformer design.

\subsection{Visualization}
\begin{figure}[H]
\centering
\includegraphics[width=0.9\linewidth]{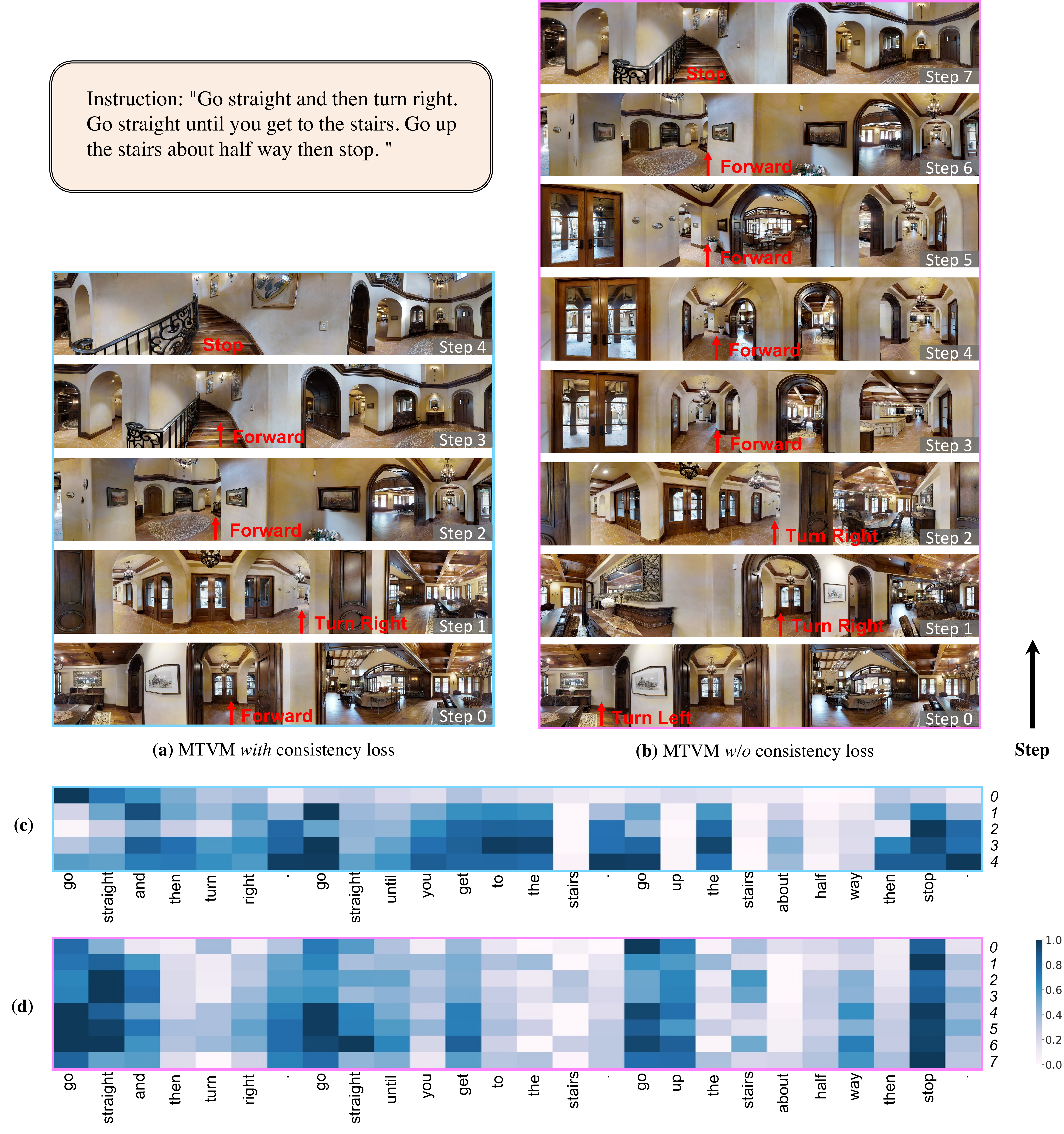}
\caption{Visualization examples of panoramic views and language attention weights.
From sub-figures (a) and (b), it can be seen that without the consistency loss, MTVM took a longer path to reach ``stairs''. 
(c) and (d) are the language attention weights at the final layer of the cross-modality encoder corresponding to (a) and (b) at each step.
%Memorized Transformer with consistency loss (c) attends from the beginning to the end of instruction and pays attention to the key words.
%In this case, the  MTVM failed to focus each step action words.
} 
\label{Vis}
\end{figure}

To demonstrate the proposed consistency loss, we give a few visualization examples of panoramic views and language attention weights in Fig.~\ref{Vis}.
In R2R dataset, the agent needs to navigate following the instruction from the beginning to the end.
Sub-figures (a) and (b) in Fig.~\ref{Vis} show that our MTVM model with the consistency loss achieves better navigation performance with a much shorter trajectory.
In sub-figures (c) and (d), we observe that our model with the consistency loss is able to better ground the sub-instructions %which shows that it can help learn the strong cross-modal representation. 
while MTVM without the consistency loss fails to focus on the action word at each step.

\section{Conclusion}
We have proposed the framework of Multimodal Transformer with Variable-length Memory (MTVM), which enables the agent explicitly model the history information in a simple and effective way. We have also designed the memory-aware consistency loss to improve the generalization ability of our model. 
Our MTVM has demonstrated strong performance, outperforming almost all the existing works on both R2R and CVDN dataset.
We see the benefit of allowing long-range dependency for VLN task and we hope this idea can benefit other vision and language interaction tasks.

\iffalse
\setlength{\tabcolsep}{4pt}
\begin{table}
\begin{center}
\caption{Font sizes of headings. Table captions should always be
positioned {\it above} the tables. The final sentence of a table
caption should end without a full stop}
\label{table:headings}
\begin{tabular}{lll}
\hline\noalign{\smallskip}
Heading level & Example & Font size and style\\
\noalign{\smallskip}
\hline
\noalign{\smallskip}
Title (centered)  & {\Large \bf Lecture Notes \dots} & 14 point, bold\\
1st-level heading & {\large \bf 1 Introduction} & 12 point, bold\\
2nd-level heading & {\bf 2.1 Printing Area} & 10 point, bold\\
3rd-level heading & {\bf Headings.} Text follows \dots & 10 point, bold
\\
4th-level heading & {\it Remark.} Text follows \dots & 10 point,
italic\\
\hline
\end{tabular}
\end{center}
\end{table}
\setlength{\tabcolsep}{1.4pt}

\begin{figure}
\centering
\includegraphics[height=6.5cm]{eijkel2}
\caption{One kernel at $x_s$ ({\it dotted kernel}) or two kernels at
$x_i$ and $x_j$ ({\it left and right}) lead to the same summed estimate
at $x_s$. This shows a figure consisting of different types of
lines. Elements of the figure described in the caption should be set in
italics,
in parentheses, as shown in this sample caption. The last
sentence of a figure caption should generally end without a full stop}
\label{fig:example}
\end{figure}

\begin{align}
  \psi (u) & = \int_{0}^{T} \left[\frac{1}{2}
  \left(\Lambda_{0}^{-1} u,u\right) + N^{\ast} (-u)\right] dt \; \\
& = 0 ?
\end{align}
\fi

\clearpage
% ---- Bibliography ----
%
% BibTeX users should specify bibliography style 'splncs04'.
% References will then be sorted and formatted in the correct style.
%
\bibliographystyle{splncs04}
\bibliography{egbib}
\end{document}